\documentclass[runningheads]{llncs}
\usepackage{graphicx}
\usepackage{hyperref}
\usepackage{comment}
\usepackage{amsmath,amssymb} 
\usepackage{color}

\usepackage{mwe}
\makeatletter
\DeclareRobustCommand\onedot{\futurelet\@let@token\@onedot}
\def\@onedot{\ifx\@let@token.\else.\null\fi\xspace}
\def\eg{\emph{e.g}\onedot}

\def\etc{\emph{etc}\onedot} 
 
\def\etal{\emph{et al}\onedot}
\def\hong{Hong~\etal~\cite{hong2018learning}}
\def\liu{Liu~\etal~\cite{liu2019learning}\ }
\makeatother

\newcommand{\fref}[1]{Fig.~\ref{#1}}
\newcommand{\tref}[1]{Table~\ref{#1}}

\newcommand{\aref}[1]{the supplementary materials}

\begin{document}
\pagestyle{headings}
\mainmatter
\def\ECCVSubNumber{4076}  


\title{SESAME: Semantic Editing of Scenes by Adding, Manipulating or Erasing Objects} 

\titlerunning{SESAME}
%
\author{Evangelos Ntavelis\inst{1,2} \and
Andr\'es Romero\inst{1} \and
Iason Kastanis\inst{2} \and
Luc Van Gool\inst{1,3} \and
Radu Timofte\inst{1}}
%
\authorrunning{E. Ntavelis et al.}

\institute{Computer Vision Lab, ETH Zurich, Switzerland \and
Robotics and Machine Learning, CSEM SA, Switzerland \and
PSI, ESAT, KU Leuven, Belgium}
\maketitle

\begin{figure}
    \centering
    \includegraphics[width=0.95\textwidth]{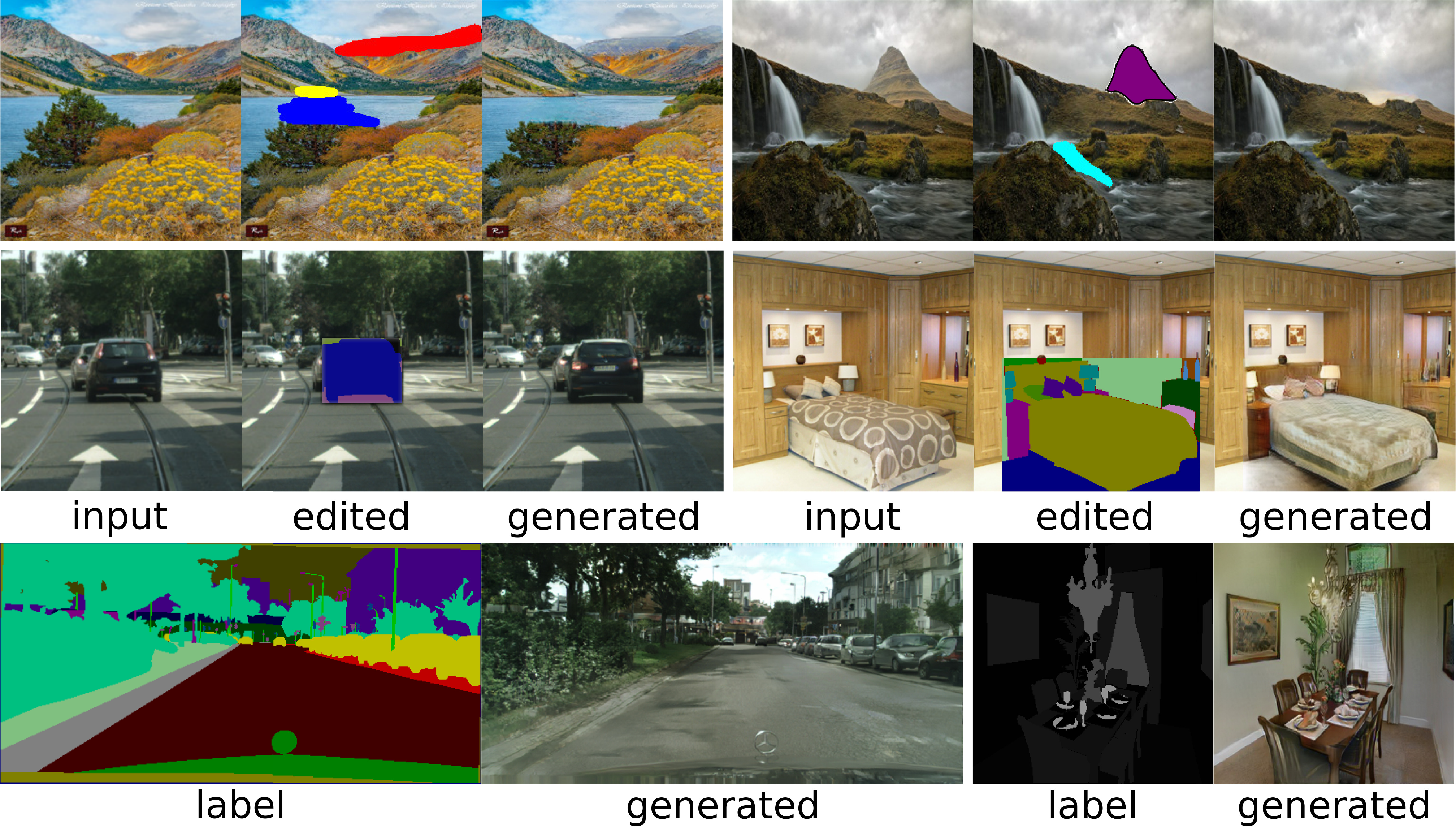}
    \caption{We assess SESAME on three tasks (a) image editing with free form semantic drawings (first row) (b) semantic layout driven semantic editing (second row) (c) layout to image generation with SESAME discriminator (third row)}
    \label{fig:teaser}
\end{figure}

\begin{abstract}
Recent advances in image generation gave rise to powerful tools for semantic image editing. However, existing approaches can either operate on a single image or require an abundance of additional information. They are not capable of handling the complete set of editing operations, that is addition, manipulation or removal of semantic concepts.
To address these limitations, 
we propose SESAME, a novel generator-discriminator pair for \textbf{S}emantic \textbf{E}diting of \textbf{S}cenes by \textbf{A}dding, \textbf{M}anipulating or \textbf{E}rasing objects.
In our setup, the user provides the semantic labels of the areas to be edited and the generator synthesizes the corresponding pixels.
In contrast to previous methods that employ a discriminator that trivially concatenates semantics and image as an input, the SESAME discriminator is composed of two input streams that independently process the image and its semantics, using the latter to manipulate the results of the former.
We evaluate our model on a diverse set of datasets and report state-of-the-art performance on two tasks: (a) image manipulation and (b) image generation conditioned on semantic labels.

\end{abstract}


\section{Introduction}
\label{sec:introduction}

Image editing is a challenging task that has received increasing attention in the media, movies and social networks.
Since the early 90s, tools like Gimp~\cite{GIMP:2018} and Photoshop~\cite{Photoshop:2020} have been extensively utilized for this task.
Yet, both require high level expertise and are labour intensive.
Generative Adversarial Networks (GANs)~\cite{goodfellowGAN} provide a learning-based alternative able to assist non-experts to express their creativity when retouching photographs.
GANs have been able to produce results of high photo-realistic quality~\cite{karras2018progressive,Karras_2019_CVPR}. Despite their success in image synthesis, their applicability on image editing is still not fully explored.
Being able to manipulate images is a crucial task for many applications such as autonomous driving~\cite{janai2017computer} and industrial imaging \cite{vidi}, where data augmentation boosts the generalization capabilities of  neural networks \cite{antoniou2017data,Wang_2018,Frid_Adar_2018}.

Image manipulation has been used in the literature to refer to various tasks.
In this paper, we follow the formulation of Bau~\etal~\cite{Bau_Ganpaint_2019}, and define the task of semantic image editing as the process of adding, altering, and removing instances of certain classes or \emph{semantic concepts} in a scene.
Examples of such manipulations include but are not limited to: removing a car from a road scene, changing the size of the eyes of a person, adding clouds in the sky, etc.
We use the term \emph{semantic concepts} to refer to various class labels that can not be identified as objects, \eg{} mountains, grass, etc.

Training neural networks for visual editing is not a trivial task.
It requires a high level of understanding of the scene, the objects, and their interconnections~\cite{shetty2018context}.
Any region of an image added or removed should look realistic and should also fit harmoniously with the rest of the scene.
In contrast to image generation, the co-existence of real and fake pixels makes the fake pixels more detectable, as the network cannot take the "easy route" of generating simple textures and shapes or even omit a whole class of objects~\cite{bau2019seeing}.
Moreover, the lack of natural image datasets, where a scene is captured with and without an object, makes it difficult to train such models in a supervised manner.

One way to circumvent this problem is by inpainting the regions of an image we seek to edit.
Following this scheme, we mask out and remove all the pixels we want to manipulate.
Recent works~\cite{yu2018free,nazeri2019edgeconnect,faceshop,Jo_2019_ICCV} improve upon this approach by incorporating sketch and color inputs to further guide the generation of the missing areas and thus provide higher level control.
However, inpainting can only tackle some aspects of semantic editing.
To address this limitation, \hong{} manipulate the semantic layout of an image and subsequently, they utilize for inpainting the image.
Yet, this approach requires access to the full semantic information of the image, which is costly to acquire.

To this end, we propose SESAME, a novel semantic editing architecture based on adversarial learning, able to manipulate images based on a semantic input.
In particular, our method is able to edit images with pixel-level guidance of semantic labels, permitting full control over the output. We propose using the semantics \emph{only} for regions to be edited, which is more cost-efficient and produces better results in certain scenarios.
Moreover, we introduce a new approach for semantics-conditioned discrimination, by utilizing two independent streams to process the input image and the corresponding semantics. 
We use the output of the semantics stream to adjust the output of the image stream.
We employ visual results along with quantitative analysis and a human study to validate the performance and flexibility of the proposed approach.

\section{Related Work}
\label{sec:related_work}

Generative Adversarial Networks~\cite{goodfellowGAN} have completely revolutionized a great variety of computer vision tasks such as image generation~\cite{Karras_2019_CVPR,karras2018progressive,miyato2018spectral}, super resolution~\cite{wang2018esrgan,lugmayr2019aim}, image attribute manipulation~\cite{mao2019mode,romero2019smit} and image editing~\cite{hong2018learning,Bau_Ganpaint_2019}.

Initially, GANs were only capable of generating samples drawn from a random distribution~\cite{goodfellowGAN}, but soon multiple models emerged able to perform \textbf{conditional image synthesis}~\cite{mirza2014conditional,odena2016conditional}.
These approaches condition the generation on various types of information.
For example, ~\cite{mirza2014conditional,miyato2018cgans,Han18,brock2018large} synthesize images characterized by a single label.
In a different setting, ~\cite{reed2016generative,han2017stackgan,stackgan++,attngan} employ a text to image pipeline to create an image based on a high-level description.
Recently, many methods utilize information of a scene graph~\cite{Johnson_2018_CVPR,Ashual_2019_ICCV} and sketches with color~\cite{sangkloy2016scribbler} to represent where objects should be positioned on the output image. 
A more fine-grained approach aims to translate semantic maps, which carry pixel-wise information, to realistic looking images~\cite{isola2017image,wang2018pix2pixHD,park2019SPADE,lee2019maskgan}. For all the aforementioned models, the user can control the output image by altering the conditional information. Nonetheless, they are not suitable for manipulating an existing image, as they do not consider an image as an input.

In \textbf{user-guided semantic image editing} the user is able to semantically edit an image by adding, manipulating, or removing semantic concepts~\cite{Bau_Ganpaint_2019}.
Both GANPaint~\cite{Bau_Ganpaint_2019} and SinGAN~\cite{Shaham_2019_ICCV} are able to perform such operations: GANPaint~\cite{Bau_Ganpaint_2019} by manipulating the neuron activations and SinGAN~\cite{Shaham_2019_ICCV} by learning the internal batch statistics of an image.
However, both are trained on a single image and require retraining in order to be applied to another, while our model is able to handle the manipulation of multiple images without retraining. 

Another simple type of editing is inpainting~\cite{iizuka2017globally,yu2018generative,liu2018image}, where the user masks a region of the image for removal and the network fills it accordingly to the image context. 
In its classic form the user does not have control over the generated pixel.
To address this, other research works guide the generation of the missing areas using edges~\cite{yu2018free,nazeri2019edgeconnect} and/or color~\cite{faceshop,Jo_2019_ICCV} information.

Recently, semantic aware approaches for inpainting address object addition and removal.
Shetty~\etal~\cite{shetty2018objectremoval} proposes a two-stage architecture to facilitate removal operations, with an auxiliary network predicting the objects' masks during training; at inference users provide them.
Note that their model cannot handle object generation.
Works in the object synthesis task are utilizing semantic layout information, a fine-grained guidance over the manipulation of an image. 
Yet, a subset of them is limited by generating objects from a single class~\cite{ouyang2018pedestrian,Wu_2019_ICCV} or placing prior fixed objects on the semantics plane~\cite{lee2018objectplacement}.
\hong{} are able to handle both addition and removal, but require full semantic information of the scene to produce even the smallest change to an image. In contrast, our method requires only the semantics of the region to be edited.

The core of the majority of the aforementioned works rely on adjusting the generator for the task of image editing. 
Most recent models use a PatchGAN variant~\cite{isola2017image} which is able to discriminate on the high frequencies of the image. This is a desired attribute as conventional losses like \textit{Mean Squared Error} and \textit{Mean Absolute Error} can only convey information about the lower frequencies to the generator. 
PatchGAN can also be used for conditional generation of images on semantic maps, similar to our case study.
Previous works targeting a similar problem concatenate the semantic information to the image and use it as an input to the discriminator.
However, conventional conditional generation literature suggests that concatenation is not the optimal approach for conditional discrimination~\cite{reed2016generative,odena2016conditional,miyato2018cgans}.
To address this, \liu{} propose a feature pyramid semantics-embedding (FPSE) discriminator using an Encoder-Decoder architecture. Each upsampling layer outputs two per-patch score maps, one trying to measure the \textit{realness} and one to gauge the \textit{semantic matching} with the labels; the later is derived after a patch-wise inner product operation with the down-sampled semantic embeddings.
Rather than incorporating a semantics loss, we use semantics to guide the image discrimination. Our model incorporates conditional information by processing it separately from the image input. In a later stage of the network, the two processed streams are merged to produce the final output of the discriminator. 

\section{SESAME}
\label{sec:proposed_approach}

\begin{figure}[t]
\begin{center}
    \centering
    \includegraphics[width=0.85\textwidth]{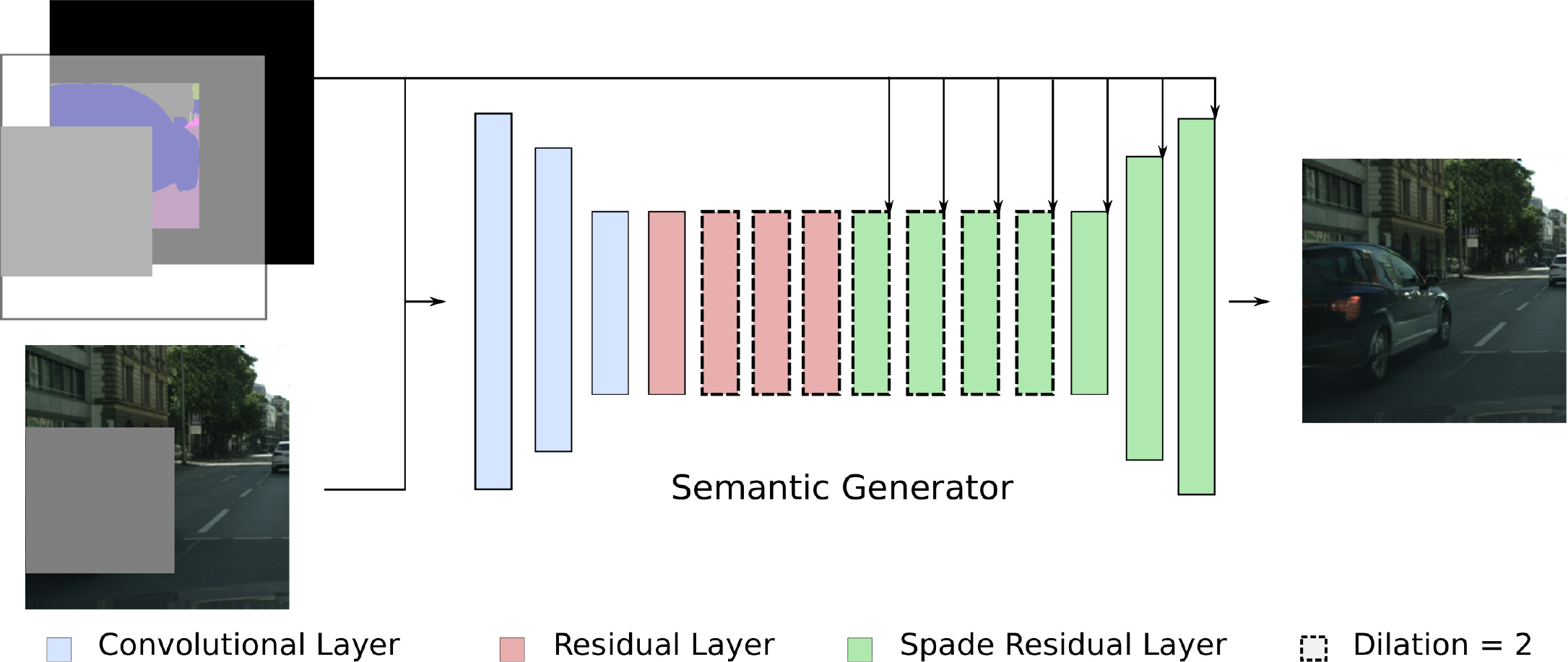}
    \caption{The SESAME Generator aims to generate the pixels designated by the semantic mask so they are both (1) true to their label and (2) fit naturally to the rest of the picture.
    It is an encoder-decoder architecture with dilated convolutions to increase the receptive field as well as SPADE layers in the decoder to guide in-class generation}
    \label{fig:generator}
\end{center}
\end{figure}

In this work we describe a deep learning pipeline for semantically editing images, using conditional Generative Adversarial Networks (cGANs). 
 Given an image $I_{real}$ and a semantic guideline of the regions that should be altered by the network, denoted by $M_{sem}$, we want to produce a realistic output $I_{out}$. 
The real pixels values corresponding to $M_{sem}$ are removed from the input image. The generated pixels in their place should be both true to the semantics dictated by the mask and coherent with the rest of the pixels of $I_{real}$.
In order to achieve this, our network is trained end-to-end in an adversarial manner. The generator is a Encoder-Decoder architecture, with dilated convolutions~\cite{Yu2016} and SPADE~\cite{park2019SPADE} layers and the discriminator is a two stream patch discriminator.

\subsubsection{SESAME Generator.} 
\label{ss:generator}

Semantically editing a scene is an Image to Image translation problem. 
We want to transform an image where we substituted some of the RGB pixels with an one-hot semantics vector. From the generator's output, only the pixels on the masked out regions are retained, while the rest are retrieved from the original image:
\begin{equation}
 I_{gen} = G(I_m, M, M_{sem}),
\end{equation}
\begin{equation}
 I_{out} = I_{gen} \cdot M + I_{real} \cdot (1-M).
\end{equation}

This architecture has two goals:  generated pixels should 1) be coherent with their real neighboring ones as well as 2) be true to the semantic input.
To achieve these goals we adapt our generator from the network proposed by Johnson~\etal~\cite{Johnson2016Perceptual} to fill the gaps: two downsampling layers, a semantic core made of multiple  residual layers and two upsampling ones. 

We conceptually divide our architecture into an encoder and a decoder part. The first extracts the contextual information of the pixels we want to synthesize. 
The seconds combines the semantic information using Spatially Adaptive De-Normalization~\cite{park2019SPADE} blocks to every layer. As the area to be edited can span over a large region, we would like the receptive field of our network to be relatively large. Thus, we use dilated convolutions in the last and first layers of the encoder and the decoder, respectively. A scheme of our SESAME generator can be seen in the \fref{fig:generator}, and for further details refer to \aref{}.

\subsubsection{SESAME Discriminator.}
\label{ss:discriminator}

\begin{figure}[t]
\begin{center}
    \centering
    \includegraphics[width=0.85\textwidth]{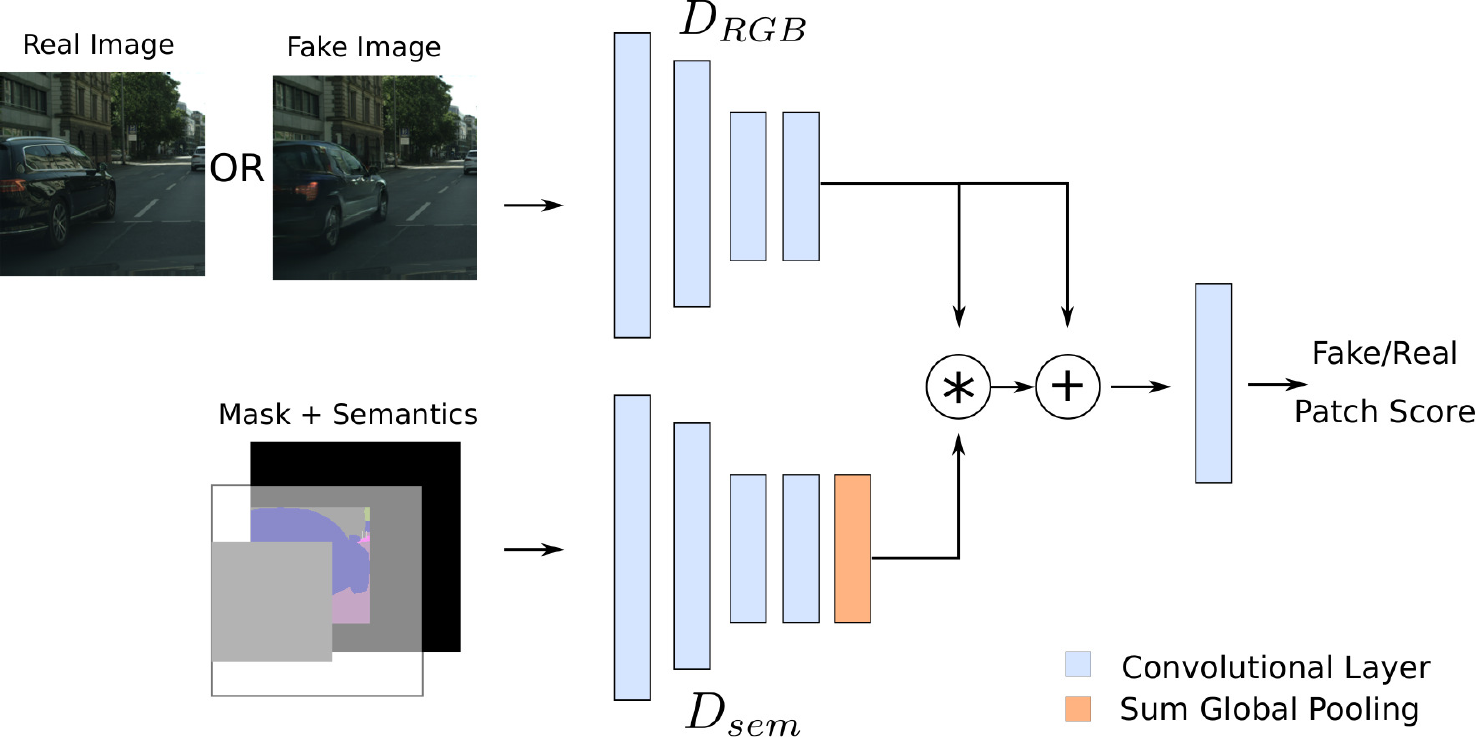}
    \caption{The SESAME discriminator, in contrast to the commonly used PatchGAN, is handling the RGB Image and its Semantics independently.
    Before the last convolutional layer the two streams, $D_{RGB}$ and $D_{Sem}$, are merged. The semantics stream is reduced via a \textit{Sum Global Pooling} operation to a 2D matrix of spatial dimensions equal to the number of output patches.
    The feature vector of $D_{RGB}$ at each path is scaled by $D_{Sem}$ and a residual is added to product
    }
    \label{fig:discriminator}
\end{center}
\end{figure}
Layout to image editing can be seen as a sub-task of label to image translation.
Inspired by the Pix2Pix~\cite{isola2017image}, more recent approaches~\cite{wang2018pix2pixHD,park2019SPADE} employ a variation of the PatchGAN discriminator.
The Markovian discriminator, as it is also called, was a paradigm shift that made the discriminator focus on the higher frequencies by limiting the attention of the discriminator into local patches, producing a different fake/real prediction value for each of them.
The subsequent methods added a multiscale discrimination approach, the Feature Matching-Loss~\cite{wang2018pix2pixHD} and the use of Spectral Normalization~\cite{miyato2018spectral} instead of Instance Normalization~\cite{park2019SPADE}, which stabilized training and further improved the quality of the generated samples.
However, the way that the conditional information was provided to the discriminator remained unchanged.

Label to image generation is a sub-task of conditional image generation. In this more general category of methods, the discriminator has evolved from the cGAN's input concatenation~\cite{mirza2014conditional}, to concatenating the class information with a hidden layer ~\cite{reed2016generative}, and lastly, to take the form of the projection discriminator~\cite{miyato2018cgans}. In the latter approach, the inner product of the embedding of the conditional information and a feature extracted from the hidden layers of the discriminator are summed with the output of the discriminator to produce the final prediction. Each step of the conditional discriminator evolution improved the results over the naive concatenation at its input~\cite{miyato2018cgans}. On all aforementioned methods the discriminator produces, nonetheless, a scalar output for the whole image.

We aim to design a discriminator for label to image generation that combines the aforementioned attributes. On the one hand, it should preserve the ability of PatchGAN to discriminate on high-frequencies. On the other hand, we want to enforce the semantic information guidance on the discriminator's decision. If the pixels of the whole image shared semantic class, the projection discriminator would be easily extended to PatchGAN. In contrast, our case is characterized by fine-grained per pixel semantics: each output patch encompasses a variety of classes and different compositions of them.

Our proposed SESAME discriminator is comprised by two independent streams that handle the RGB and Semantic Labels inputs.
As \fref{fig:discriminator} depicts, the two streams have identical architectures. 
Before the information is merged a \textit{Sum Global Pooling} operation is applied to the output of the Semantic Stream.
The output of the semantic stream is used to scale each output coming from the RGB stream.
The resulted feature map is passed as input to a last $3\times3$ convolutional layer, which produces the final output.
The process can be written as follows:
\begin{equation}
   D(I,Sem) = {Conv_{3\times3}}( \thinspace D_{RGB}(I_{out}) \cdot(1 + \sum_{channels}{D_{sem}(Sem)}) \thinspace ),
\end{equation}\label{eq:scale}
where the $D_{RGB}$ is the output of the RGB stream and $D_{sem}$ of the semantic stream before the \textit{Global Sum Pooling}.
We also integrate the changes made to PatchGAN by Pix2PixHD~\cite{wang2018pix2pixHD} and SPADE~\cite{park2019SPADE}.
We use a multiscale discrimination scheme with squared patches and two different edge-sizes of 70 and 140 pixels, in order to provide also discrimination at a coarser level and Spectral Normalization. 
The input to the semantic stream is the same for both fake and real images discrimination, so we only need to calculate $D_{sem}$ once. 
Moreover, it makes sense to apply the Feature Matching Loss only to the Feature Maps produced by the RGB stream.

\subsubsection{Training Losses.}
We train the Generator in an adversarial manner using the following losses: Perceptual Loss~\cite{Johnson2016Perceptual}, Feature Matching Loss~\cite{NIPS2016_6125}, and Hinge Loss~\cite{lim2017geometric,tranNIPS2017_7136,miyato2018spectral} as the Adversarial Loss.
Early experiments with Style Loss~\cite{Johnson2016Perceptual} did not improve the results.
Accordingly:
\begin{equation}
L_G = \lambda_{percept} \cdot L_{perc} + \lambda_{feat} \cdot L_{FM} 
           - E_{z\sim p(z)}[D_k(I_{out},M,M_{sem}))],
\end{equation}
Where each $\lambda$ represents the relative importance of each loss component.

For the discriminator at each scale, the Hinge Loss takes the following form: 
\begin{multline}
L_{D_k} = E_{z\sim q_{data}(x)}[min(0, -1 + D_k(I_{real},M,M_{sem}))] + \\
E_{z\sim p(z)}[min(1, -1 - D_k(I_{real},M,M_{sem}))],
\end{multline}
which is then combined to form the full discrimination loss,
\begin{equation}
 L_D = L_{D_1} + L_{D_2}.
\end{equation}
\label{ss:preparation}

\section{Experiments}
\label{sec:experiments}

In Section \ref{sec:proposed_approach} we described how the SESAME Generator can be used to semantic edit images for addition, manipulation, and removal, and how we designed the SESAME discriminator to tackle both image editing and generation. To elucidate the merits of our approach, we conducted a series of different experiments:
\begin{itemize}
    \item In order to quantify the performance of our network we follow the data preparation and evaluation steps of \hong{}, for generating and removing objects based on a given semantic layout.
    \item We train our model to permit free form semantic input from users to manipulate scenes and show qualitative results.
    \item We train our SESAME discriminator along with SPADE Generator for Label to Image Generation.
\end{itemize}

\begin{figure}[t]
\begin{center}
    \centering
    \includegraphics[width=1\textwidth]{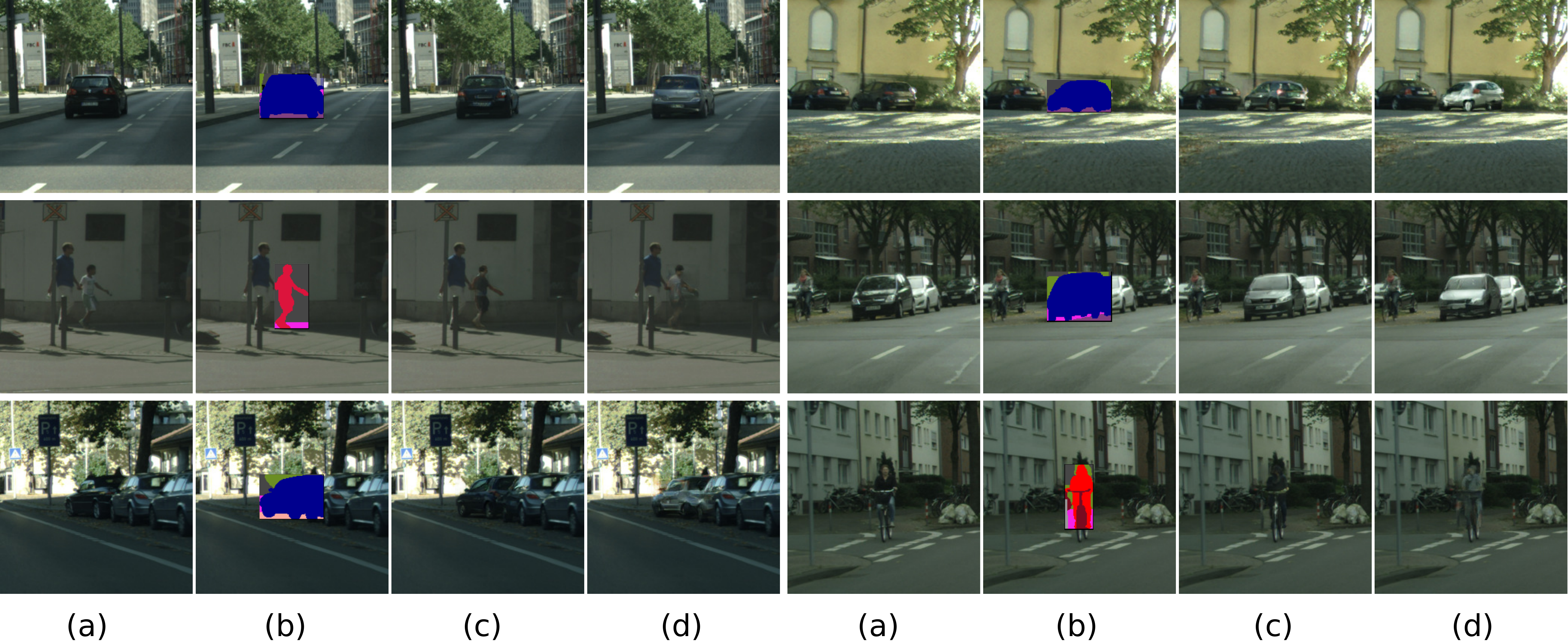}
    \caption{Visual results of addition on Cityscapes: (a) input image (b) edited semantics (c) SESAME (d)\hong{}. Note that \hong{} require the whole semantics while we use only the semantics of the box}
    \label{fig:city_results}
\end{center}
\end{figure}

\begin{figure}[t]
\begin{center}
    \centering
    \includegraphics[width=1\textwidth]{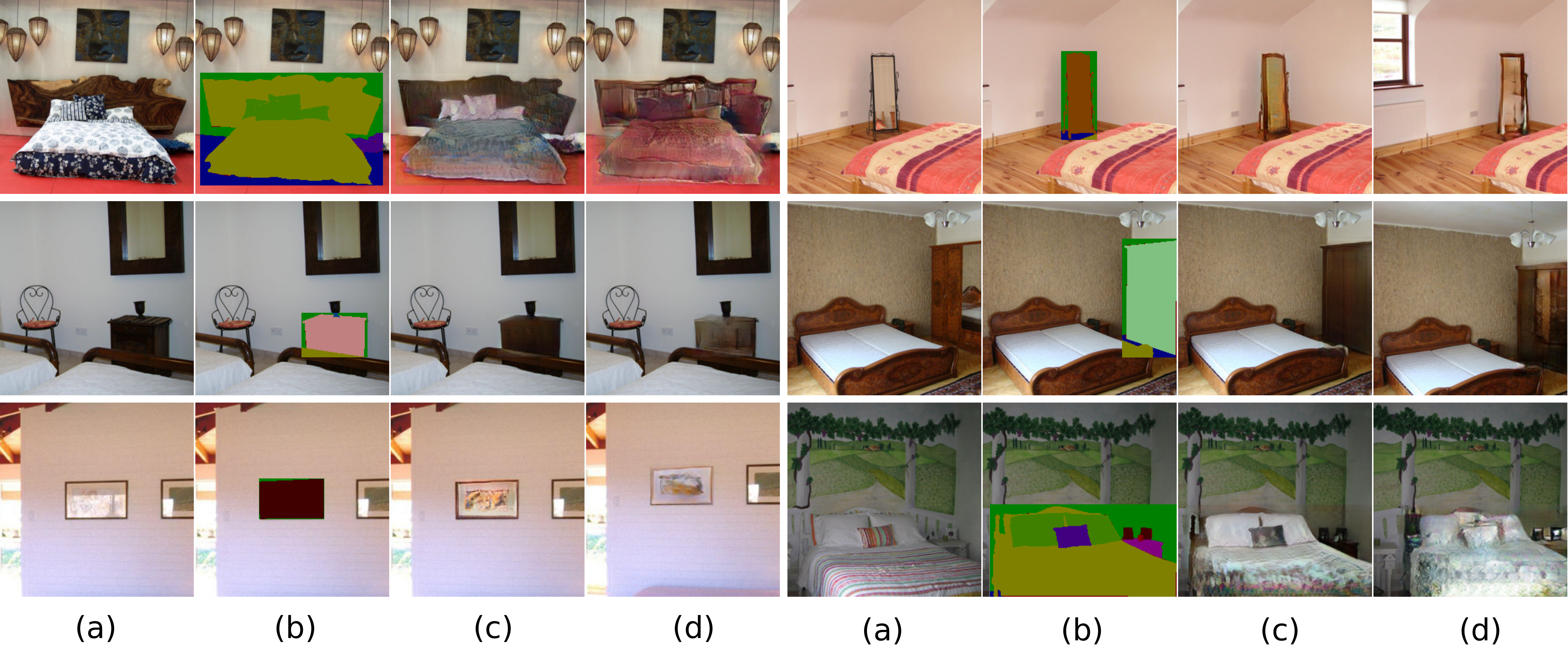}
    \caption{Visual results of addition on ADE20k: (a) input image (b) edited semantics (c) SESAME (d) \hong{}. In this setting we use the full semantic information to guide the editing}
    \label{fig:ade_results}
\end{center}
\end{figure}

\begin{figure}[t]
\begin{center}
    \centering
    \includegraphics[width=1\textwidth]{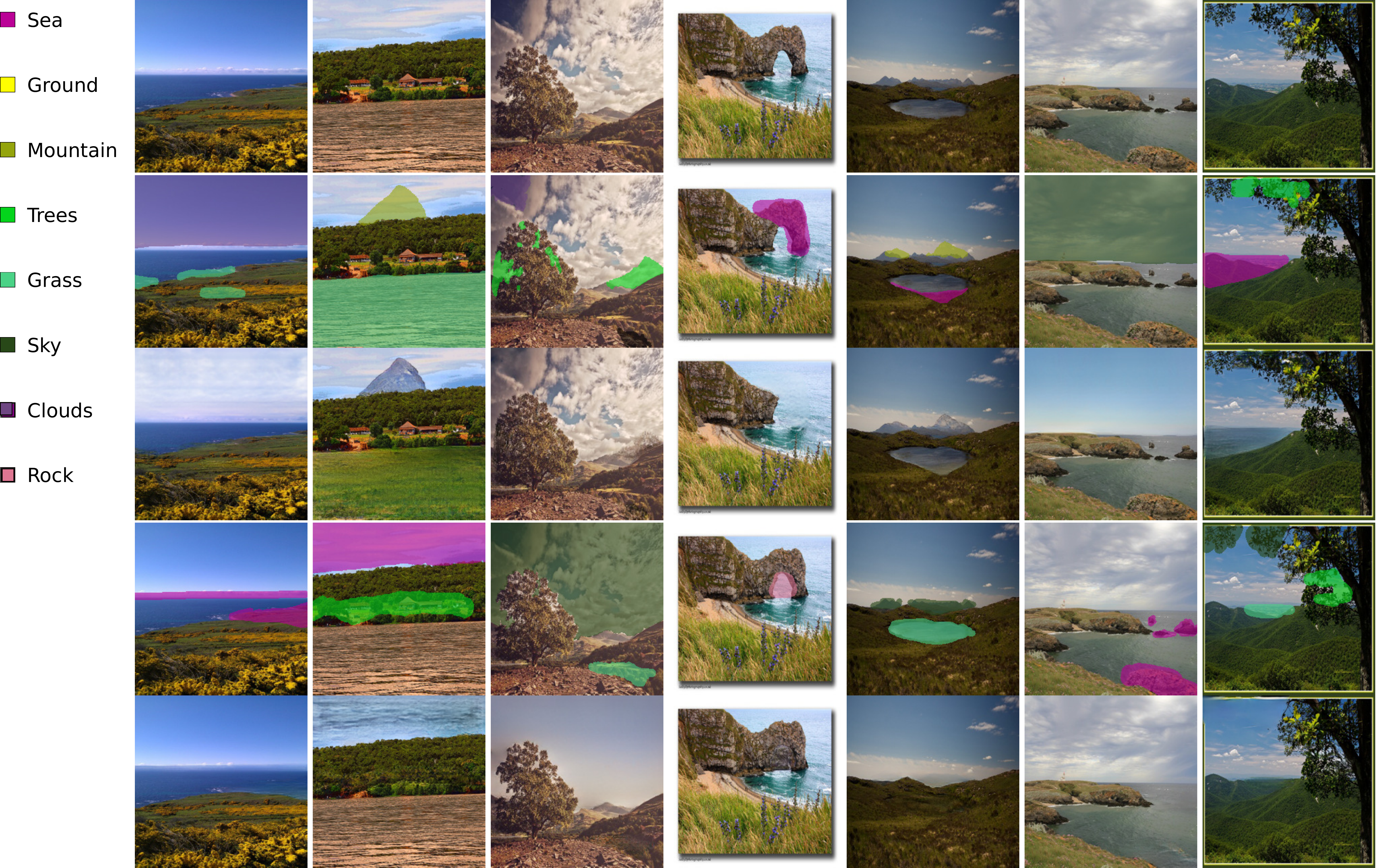}
    \caption{Free-form image manipulation. The user can select a semantic brush and paint over the image to adjust as they see fit}
    \label{fig:freeform}
\end{center}
\end{figure}


\subsubsection{Implementation Details.}
\label{ssc:implementation_details}
For training we are using the Two Time-Scale Update Rule~\cite{NIPS2017_7240} to determine the scale between the learning rate of the generator and the discriminators, with $lr_{gen} = 0.0001 $ and $lr_{disc} = 0.0004$.
We train for 200 epochs.
After 100 we start to linearly decay the learning rates to $0$.
For our generator losses we multiplied the Feature Matching Loss and Perceptual loss by a factor of $10$ before adding them to the adversarial loss. We use the Adam optimizer~\cite{adam} with coefficient values of $b_1 = 0$ and $b_2 = 0.999$, similar to~\cite{park2019SPADE}.


\subsubsection{Datasets.} In line with the literature we conduct experiments on:
\begin{itemize}
    \item\textbf{Cityscapes~\cite{Cordts_2016_CVPR}}. The dataset contains 3,000 street-level view images of 50 different cities in Europe for the training set and 500 images for the validation set. The images are accompanied by fine-grained information of the per-pixel semantics and instance segmentation with original resolution of the images is $2048\times 1024$ pixels. For addition and removal, we downsample to $1024\times 512$ pixels before patches of $256\times 256$ pixels are extracted.
    Following \hong{}, we choose 10 of the 30 available semantics classes as foreground objects, \eg{} \textit{pedestians, cars, bicycles}, \etc{}.
    For generation we resize the image to $512\times 256$ pixels.
    
    \item\textbf{ADE20K~\cite{zhou2016semantic,zhou2017scene}}
    \label{ssc:ADE20K}
    ADE20K has over 20,000 images together with their detailed semantics for 150 different semantic classes.
    In addition, 2,000 more images are offered for validation.
    The whole dataset is used for the generation task.
	For manipulation, following Hong~\etal~\cite{hong2018learning}, we experiment on a subset of the ADE20K dataset comprised of bedroom scenes.
	Similarly to Cityscapes, 31 objects are chosen as foreground objects. In total we consider 49 semantic categories for training and evaluation. 
     
    \item\textbf{Flickr-Landscapes Datasets~\cite{park2019SPADE}}
    \label{ssc:Flickr}
    Similar to SPADE~\cite{park2019SPADE}, we first scrapped 200,000 images from flickr with only landscape constraint. As our main purpose is to show image editing over significant areas within a landscape, we use a DeepLab-v2~\cite{chen2017deeplabv2} network trained on COCO-Stuff in order to extract images that contain at least 80\% pixels of clouds, mountains, water, grass, etc. After post-processing, our curated dataset consists of 7367 training and 500 validation images with their corresponding segmentation for 17 different semantic classes. 
\end{itemize}

\subsubsection{Data Preprocessing.}
Free-form semantic editing is not trivial to achieve.
The model can easily overfit on mask shapes used during training.
In order to train our free-form semantic editing experiments, we randomly draw a box mask in conjunction with random strokes~\cite{yu2018free} with $70\%$ chance, otherwise we drop all the pixels belonging to a semantic class of the training image.
   
For layout driven editing, we extend the data pre-processing scheme introduced by Hong~\etal~\cite{hong2018learning}.
A rectangular area is removed from the input image and we try to inpaint it using the semantic labels.
To train the addition operation, they extract the boundary boxes based on the instances of the foreground classes.
For the removal subtask, they randomly choose and remove blocks to train the network to inpaint background classes.
While this makes sense for a dataset like ADE20k where the foreground objects can be found anywhere in the pictures, in Cityscapes the foreground objects placement follow certain distributions~\cite{lee2018objectplacement}.
Thus, we only extract a randomly chosen rectangular area if it contains at least a pixel of \textit{ground, road, sidewalk and parking}.

\subsubsection{Quantitative Results.}
\label{ssc:quantitative_results}

\begin{table}[t]
\begin{center}
\caption{Addition Results for Cityscapes and ADE20k dataset. We ablate on the Generator and Discriminator architecture as well as the semantic availability. For the SSIM, accuracy and mIoU higher is better, while for FID, lower is better.}
\label{table:addition}
\begin{tabular}{c|c|c|cccc|cccc}
\multicolumn{3}{c}{} &  \multicolumn{4}{c}{Cityscapes} &  \multicolumn{4}{c}{ADE20k} \\
\cline{4-11}
Generator & Disc & Labels & SSIM & accu  & mIoU  & FID  & SSIM  & accu  & mIoU   & FID  \\
\hline
\hong{} & PatchGAN & Full %
 & 0.377 & 83.8 & 60.7 & 12.11 %
 & 0.205 & 92.2 & 34.6 & 28.48 \\
\hong{} & PatchGAN & BBox %
& 0.379 & 85.9 & 63.4 & 11.50 %
& 0.183 & 92.7 & 35.3 & 28.36 \\
\hong{} & SESAME & Full %
& 0.396 & 86.0 & 64.0 & 11.76 %
& 0.192 & 91.3 & 34.1 & 28.55 \\
SESAME & PatchGAN & BBox %
& 0.375 & 86.0 & 64.5 & 11.13 %
& 0.193 & 92.2 & 35.7 & 28.16 \\
SESAME & SESAME & BBox %
& \textbf{0.410}  & \textbf{86.0} & \textbf{65.3}  & \textbf{11.03} %
& \textbf{0.209} & \textbf{93.3} & \textbf{37.1} & \textbf{26.66} \\
\end{tabular}
\end{center}
\end{table}

\begin{table}[t]
\begin{center}
\caption{Removal results for Cityscapes and ADE20k datasets. For the SSIM, accuracy and mIoU higher is better, while for FID, lower is better.}
\label{table:removal}
\begin{tabular}{l|cccc|cccc}
\multicolumn{1}{l}{} &  \multicolumn{4}{c}{Cityscapes} & \multicolumn{4}{c}{ADE20k} \\
\cline{2-9}
Method & SSIM$\uparrow$ & accu$\uparrow$  & mIoU$\uparrow$  & FID$\downarrow$  & SSIM$\uparrow$  & accu$\uparrow$  & mIoU$\uparrow$   & FID$\downarrow$  \\
\hline
\hong{} %
& 0.584 & 83.9\% & 65.3\% & 10.34 %
&  0.456  & 91.7\% & 40.0\% &  24.98  \\
SESAME %
& \textbf{0.797} & \textbf{85.0}\% & \textbf{67.6}\% & \textbf{7.43}  %
& \textbf{0.491} & \textbf{92.3\%} & \textbf{41.6\%} &  \textbf{23.30} \\
\end{tabular}
\end{center}

\end{table}

For measuring the performance of our network, we combine the evaluation approach of previous methods~\cite{hong2018learning,park2019SPADE}.
To assess the perceptual quality of our synthesis we use the Frechet Inception Distance (FID)~\cite{NIPS2017_7240}.
We compare the mean Intersection over Union (mIoU), and the pixel-accuracy loss between the ground truth semantic layout and the inferred one. We infer the semantic labels of the generated images using pretrained semantic segmentation networks~\cite{Yu2016,Yu2017,zhou2016semantic,zhou2017scene}.
In order to maintain consistency, we chose the same object for validating all our experiments.
Additionally, in our comparison with \hong{}, 
we compute the Structural Similarity Index (SSIM)~\cite{wang2004image} between each $\langle I_{real}, I_{out} \rangle$ pair, taking into account only the generated pixels.
Naturally, as in the editing task only a small percentage of image pixels are changed we expect better results than those in the Generation experiments, but also better methods yield larger performance gains when tackling the latter.

\subsubsection{Our Baselines.}
\label{ssc:baseline}

For our semantic image editing baseline we are using the work of Hong~\etal~\cite{hong2018learning}. 
They introduced a hierarchical model to tackle the task of image editing. On the first stage, they inpaint the semantic classes of an image with a missing region. Then they combine the predicted output with the ground truth and after concatenating the real image with the missing pixels, they use their second stage model to fill the image.
Similar to their work, we focus on the \textit{mask to image generation} task and compare our model against their image generator trained on the ground truth labels.
Their approach consists of an encoder and a decoder.
The encoder has two input streams where the image and the semantics are processed separately and are then \textit{fused} based on the mask of the object location.
The result of the fusion is then passed to an image decoder which produces the end result. The generator is trained in conjunction with a PatchGAN discriminator.
We use different architectures and largely decrease the number of parameters for the generator and have a larger discriminator as shown in Table~\ref{table:parameters}. However, during inference time only the generator is used. Reduced number of parameters for the generator is clearly beneficial during execution. 
For Image Generation we compare against SPADE~\cite{park2019SPADE} and CC-FPSE~\cite{liu2019learning}.

\subsubsection{Addition and Removal of Objects} To demonstrate the ability of our network to perform well both on the addition and the removal part we compare on both tasks separately. The computed metrics for these cases can be found in Tables~\ref{table:addition} and \ref{table:removal}, respectively. 

\begin{table}[t]
\begin{center}
\caption{Comparison in number of parameters}
\begin{tabular}{l|c|c}
\multicolumn{1}{l}{} & \multicolumn{2}{c}{Parameters in millions} \\
\cline{2-3}
 Method & Generator & Discriminator \\
\hline
\hong & 190m & 5.6m \\
SESAME & 20.5m & 11.1m \\
\end{tabular}
\end{center}
\label{table:parameters}
\end{table}

In the visual results we can observe that objects look sharper and their features are more distinctive.
Furthermore, as Figures \ref{fig:city_results} and \ref{fig:ade_results} illustrate, our method generates different patterns for different clothes, and cars in which the windows are not mixed with the rest of the car. Besides our better numerical results, our user study further illustrates the superiority of our approach.
In the case of removal, artifacts of the \textit{BBox} are commonly left in picture by the method of 
\hong{}, whilst in our case this effect is difficult to notice. 

\subsubsection{Labels to Image Generation}

\begin{figure}[t]
\begin{center}
    \centering
    \includegraphics[width=1\textwidth]{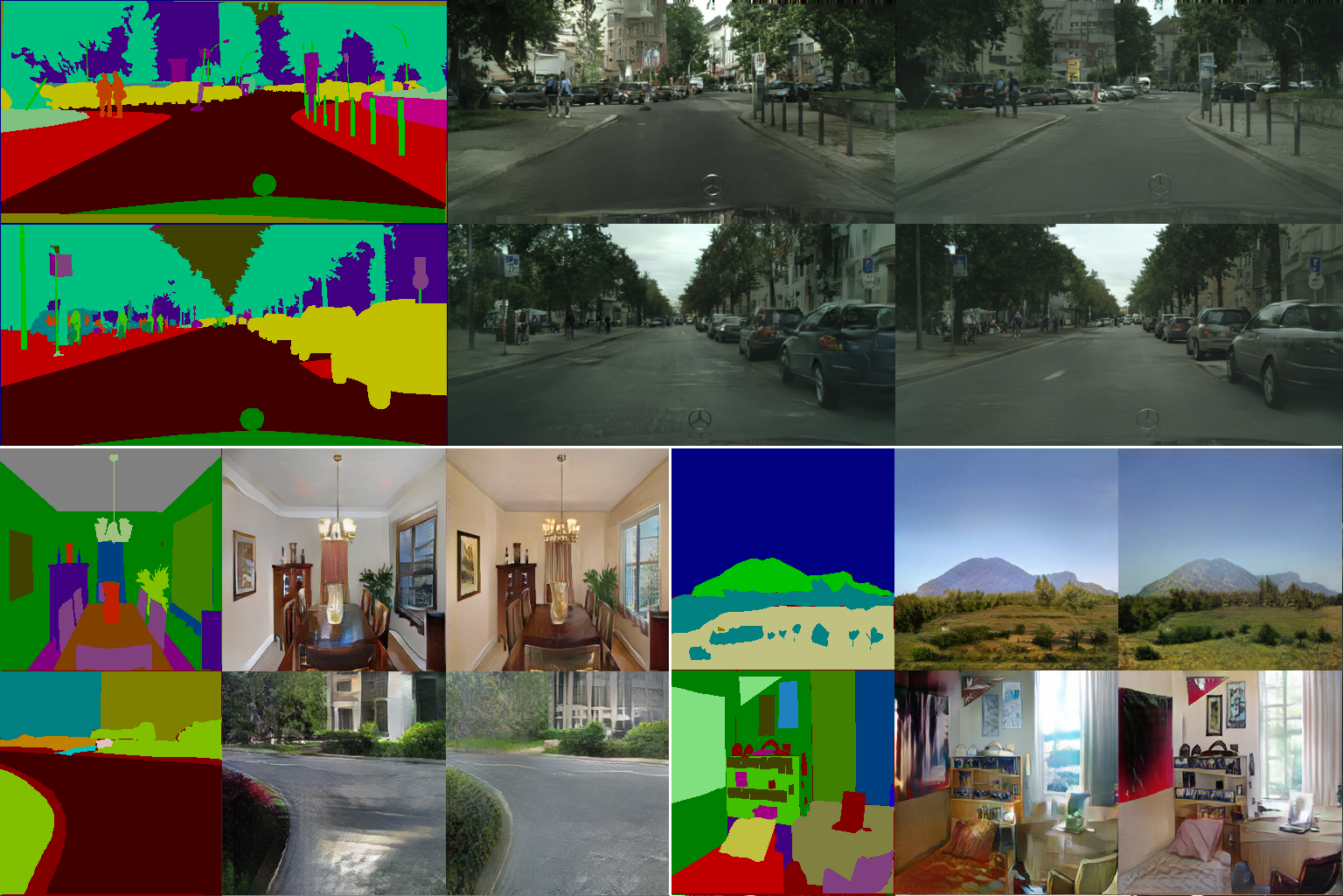}
    \caption{Label to Image Generation results. For each triplet of images we are showing the semantic layout input (left), generation using PatchGAN (center) and SESAME (right) discriminator on top of SPADE generator}
    \label{fig:generation}
\end{center}
\end{figure}

The SESAME Discriminator is designed to tackle the shortcomings of the naive concatenation of an image and its semantics label when generating images.
We measure the performance on Labels to Image Generation against SPADE~\cite{park2019SPADE} using the same generator and against CC-FPSE of \liu{}. Please refer to SM for the differences in our approaches. 
The results for Cityscapes and ADE20k datasets can be found on~\tref{table:generation} and \fref{fig:generation}.

\begin{table}[t]
\begin{center}
\caption{Layout to image generation results. For mIoU and accu, higher is better, while for FID, lower is better}
\label{table:generation}
\begin{tabular}{c|c|ccc|ccc}
\multicolumn{2}{c}{} &  \multicolumn{3}{c}{Cityscapes} &  \multicolumn{3}{c}{ADE20k} \\
\cline{3-8}
Generator & Discriminator & mIoU & accu & FID & mIoU & accu & FID \\
\hline
Pix2PixHD & PatchGAN & 58.3 & 81.4 & 95.0 & {20.3} & 69.2 & 81.8 \\
Pix2PixHD & SESAME & 59.6 & 81.1 & 55.4 & 49.0 & 85.5 & 36.8 \\
SPADE & PatchGAN & 62.3 & 81.9 & 71.8 &  38.5 & 79.9 & 33.9 \\
CC & FPSE & 65.5 & 82.3 & 54.3 &  43.7 & 82.9 & \textbf{31.7} \\
SPADE & SESAME & \textbf{66.0} & \textbf{82.5} & \textbf{54.2} & \textbf{49.0} & \textbf{85.5} & 31.9 \\

\end{tabular}
\end{center}
\end{table}

\subsubsection{Free-Form Semantic Image Editing}
\label{ss:freeform}

The user selects a brush of a semantic class and paints over the image.
The pixels that are painted over are removed from the image and SESAME is filling the gaps based on the painted semantic guidance.
Examples of hand painted masks and corresponding results can be seen on \fref{fig:freeform}.
Additionally, the context is very important for the label we want to add: a patch of grass cannot be drawn in the middle of the sky. More results can be found in \aref{}.

\begin{table}[t]
    \centering
    \caption{\textbf{User Study Results:} \textit{Which image is the most photorealistic?} The first study invited the users to choose between \hong with full semantics information and ours with full and bbox semantics, respectively. The second study invited to choose between the results produced by our SESAME and the PatchGAN discriminator, for different availabilities of semantics} 
    \label{table:user_study}    
    \begin{tabular}{c|c}
    \multicolumn{2}{c}{User Study I}\\
    \hline
    Setting & Preference [\%]\\
    \hline
    \hong{} &  22.50 \\
    SESAME w Full &  35.83  \\
    SESAME w BBox &  \textbf{41.67} \\
    \end{tabular}
    \begin{tabular}{c|cc}
      \multicolumn{3}{c}{User Study II}\\
      \hline
    Discriminator: & SESAME & PatchGAN \\
    \hline
    FullContext & \textbf{56.67} & 43.33\\
    BBoxContext & \textbf{61.04} & 38.96\\
    \end{tabular}

\end{table}

\subsubsection{Ablation Study}
\label{ssc:ablation_study}

SESAME incorporates a Generation/Discrimination pair able to edit a scene by only considering the Semantics of regions in the image that the user seeks to edit. In order to showcase the benefits of our approach we ablate the performance of our architecture by varying (a) the generator architecture, (b) the discriminator architecture for both image manipulation and generation and (c) the available semantics only for manipulation, by utilizing either the \emph{Full} semantic layout or the semantics of the rectangular region we want to edit, which we refer to as \emph{BBox} Semantics. 

As we observe in \tref{table:addition} and \tref{table:generation} in almost all cases using the SESAME discriminator improves the performance compared to the commonly used PatchGAN. Our generator is producing better visual quality results(mIoU, FID) but is lagging in fidelity(SSIM) when compared with the one from \hong{}. The partial \emph{BBox} semantics improve the performance in the case of Cityscapes dataset but full semantics work better for ADE20k. We observe the same effect when using full semantics for SESAME. 

In another series of experiments, we substituted our Semantics merging operation. Instead of applying \textit{Sum Global Pooling}, we experiment with $1)$ concatenating the two streams of information and $2)$ calculating their element-wise product resulted in lower FID score compared to the proposed approach, $11.96$ and $12.02$ respectively.

\subsubsection{User Study}
\label{ssc:user_study}

We employed Amazon Mechanical Turk\footnote{\url{https://www.mturk.com}} for the two experiments of our user study.
For each of them we took $100$ samples from our validation set and asked 20 Turkers: \textit{Which among the images looks more photorealistic?}

The first experiment presented the Turkers with three options: our method with access to only the \textit{BBox} information and both ours and \hong{} model using the \textit{Full} Semantics. 
As shown in Table~\ref{table:user_study} the results of our SESAME approach were clearly preferred by the users over the results of our baseline. Moreover, in agreement with our quantitative analysis, the proposed scaling scheme in our discriminator benefits from less irrelevant semantic information.
Another group of settings compared the results when the PatchGAN is used instead of our SESAME discriminator. The results consistently show that the independent processing of the semantic information leads to better perceptual quality of the results; they are picked more often by the human subjects.

\section{Conclusion}
\label{sec:conclusion}
In this work, we introduce SESAME a novel method for semantic image editing covering the complete spectrum of adding, manipulating, and erasing operations. 
Our generator is capable of manipulating an image by only conditioning on the semantics of the regions the user seeks to edit, namely without requiring the information about the full layout.
Our discriminator processes the semantic and image information in separate streams and overcomes the limitations of the concatenating approach inherent in PatchGAN.
SESAME produces state-of-the-art results on the tasks of (a) semantic image manipulation and (b) layout to image generation and permits the user to edit an image by intuitively painting over it. 
As a future research direction, we plan to extend this work on image generation conditioned on other types of information, \eg{} scene graphs could also benefit from our two-stream discriminator.
We refer to supplementary material for more details and to \href{https://github.com/vglsd/OpenSESAME}{OpenSESAME} for the code and the pretrained models.

\subsubsection{Acknowledgements} 
This work was partly supported by CSEM, ETH Zurich Fund (OK) and by Huawei, Amazon AWS and Nvidia GPU grants. 
We are grateful to Despoina Paschalidou, Siavash Bigdeli and Danda Pani Paudel for fruitful discussions. We also thank Gene Kogan for providing guidance on how to prepare the Flickr Landscapes Dataset. 

\bibliographystyle{splncs04}
\bibliography{egbib}

\clearpage
\appendix
\section{Supplementary Material}
\subsubsection{Model Architectures}
The detailed architectures of the SESAME generator and discriminator are depicted in Tables \ref{tab:gen_arch} and \ref{tab:dis_arch} respectively. Please note that for the discriminator we use this architecture twice, once for each scale.

To further justify the architectural choices of our generator we compare against Pix2PixHD++~\cite{wang2018pix2pixHD},  with SPADE layers on the decoder part. 
We use both our SESAME discriminator and the commonly used PatchGAN.

The performance assessment is reported in \tref{table:ablation}. The use of our discriminator over PatchGAN improved the results in almost all cases.
However, we observe that while our generator-discriminator combination performs the best, the second combination is without any of our networks. We argue that our proposed method works better together as the large receptive field provided by the dilated convolutions in our generator synergizes well with the highly focused gradient flow coming from our discriminator. 

\begin{table}[]
    \centering
    \caption{SESAME generator architecture. We depict the number of \textbf{F}ilters, the \textbf{K}ernel size, the \textbf{S}tride and the \textbf{D}ilation factor}
    \begin{tabular}{c|c|c}
    Layer & Normalization & Activation \\
    \hline
ConvBlock F = 64, K = 7, S = 1, D = 1 & Instance & ReLU \\
ConvBlock F = 128, K = 3, S = 2, D = 1 & Instance & ReLU \\
ConvBlock F = 256, K = 3 , S = 2, D = 1 & Instance & ReLU \\
ResBlock F = 256, K = 3, S = 1, D = 1 & Instance & ReLU \\
ResBlock F = 256, K = 3, S = 1, D = 2 & Instance & ReLU \\
ResBlock F = 256, K = 3, S = 1, D = 2 & Instance & ReLU \\
ResBlock F = 256, K = 3, S = 1, D = 2 & Instance & ReLU \\
ResBlock F = 256, K = 3, S = 1, D = 2 & SPADE & LeakyReLU(0.02) \\
ResBlock F = 256, K = 3, S = 1, D = 2 & SPADE & LeakyReLU(0.02) \\
ResBlock F = 256, K = 3, S = 1, D = 2 & SPADE & LeakyReLU(0.02) \\
ResBlock F = 256, K = 3, S = 1, D = 2 & SPADE & LeakyReLU(0.02) \\
ResBlock F = 256, K = 3, S = 1, D = 1 & SPADE & LeakyReLU(0.02) \\
\textit{Nearest Neighbour Upsampling $\times$2} & - & - \\
ResBlock F = 128, K = 3, S = 1, D = 1 & SPADE & LeakyReLU(0.02) \\
\textit{Nearest Neighbour Upsampling $\times$2} & - & - \\
ResBlock F = 64, K = 3, S = 1, D = 1 & SPADE & LeakyReLU(0.02) \\
ConvBlock F = 3, K = 3, S = 1, D = 1 & - & TanH \\
    \end{tabular}
    \label{tab:gen_arch}
\end{table}

\begin{table}[]
    \centering
    \caption{SESAME discriminator architecture \textbf{per scale}, We depict the number of \textbf{F}ilters, the \textbf{K}ernel size, the \textbf{S}tride and the \textbf{D}ilation factor}
        \begin{tabular}{c|c|c}
    Layer & Normalization & Activation \\
    \hline
    \multicolumn{3}{c}{Image Stream} \\
    \hline
ConvBlock F = 64, K = 4, S = 2, D = 1 & - & LeakyReLU(0.02) \\
ConvBlock F = 128, K = 4, S = 2, D = 1 & SpectralInstance & LeakyReLU(0.02) \\
ConvBlock F = 256, K = 4, S = 2, D = 1 & SpectralInstance & LeakyReLU(0.02) \\
ConvBlock F = 512, K = 4, S = 1, D = 1 & SpectralInstance & LeakyReLU(0.02) \\
    \hline
    \multicolumn{3}{c}{Semantics Stream} \\
    \hline
ConvBlock F = 64, K = 4, S = 2, D = 1 & - & LeakyReLU(0.02) \\
ConvBlock F = 128, K = 4, S = 2, D = 1 & SpectralInstance & LeakyReLU(0.02) \\
ConvBlock F = 256, K = 4, S = 2, D = 1 & SpectralInstance & LeakyReLU(0.02) \\
ConvBlock F = 512, K = 4, S = 1, D = 1 & SpectralInstance & LeakyReLU(0.02) \\
\textit{Sum Global Pooling} & - & - \\
\hline
\multicolumn{3}{c}{Common Head} \\
\hline
ConvBlock F = 1, K = 4, S = 1, D = 1 & - & - \\
    \end{tabular}
    \label{tab:dis_arch}
\end{table}

\subsubsection{Replacement of objects}
Apart from adding and removing objects, SESAME can also be used to replace an instance of an object, given that we know its class and its outline.
SESAME can be utilized in this manner for dataset augmentation.
We conduct experiments on replacing objects in street scenes of Cityscapes and we ablate on the usage of our SESAME discriminator against the PatchGAN. In Table \ref{table:replacement} we measure the FID score of the image results, the SSIM of the generated regions and we also devoted a part of our user study, described in Section 4 of the main paper, to test which of the two discriminators produces the most \emph{photo-realistic} results. Visual results can be found in Figure \ref{fig:city_add}.

\begin{table}[t]
\centering
\caption{We ablate on the semantics availability, the generator architecture and the discriminator architecture for adding objects on street scenes from Cityscapes w.r.t. FID score (lower is better) }
\label{table:ablation}
\begin{tabular}{l|cc|cc}
\multicolumn{1}{l}{} &  \multicolumn{4}{c}{Discriminator} \\
\multicolumn{1}{l}{} &  \multicolumn{2}{c|}{Full semantics} & \multicolumn{2}{c}{BBox semantics} \\
\cline{2-5}
Generator & PatchGAN & SESAME & PatchGAN & SESAME \\
\hline
SPADEPix2PixHD & 11.92 & 12.74 & 12.32 & 12.66 \\ 
SESAME & 11.95 & 11.64 & 11.13 & \textbf{11.03} \\
\end{tabular}
\end{table}

\begin{table*}
\centering
\caption{Object Replacement - Cityscapes. We show the performance of our SESAME model with our discriminator and the PatchGAN\cite{isola2017image,park2019SPADE} discriminator as well as the percentage of user answers to the question: \textit{Which image looks more photo-realistic?}}
\begin{tabular}{l|cc|c}
Discriminator & SSIM$\uparrow$ & FID$\downarrow$ & User Preference(\%) \\
\hline
PatchGAN &  0.390 & 10.63 & 45.03 \\
SESAME & \textbf{0.433} & \textbf{9.3} & \textbf{54.97} \\
\end{tabular}
\label{table:replacement}
\end{table*}


\subsubsection{Visual Results}

\begin{figure}[]
\begin{center}
    \centering
    \includegraphics[width=1\textwidth]{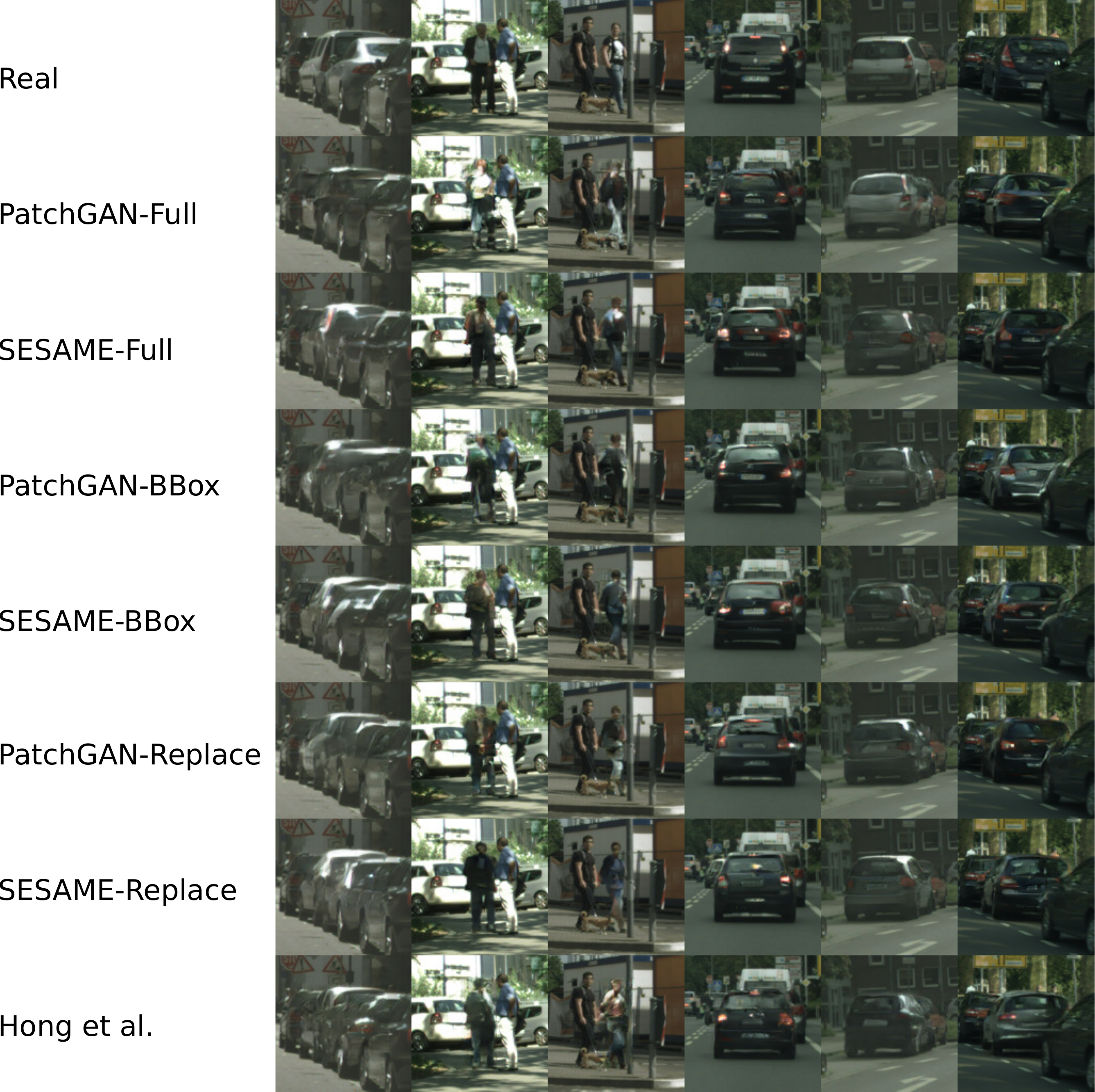}
    \caption{Visual results for addition on Cityscapes\cite{Cordts_2016_CVPR}. We ablate on: (a) using the Full context, using only the labels of the rectangular areas to be edited and only replacing an object given its mask (b) generations due to training with the PatchGAN Discriminator and SESAME. Finally, we show the results produced by the method of \hong{}, using the full semantics information}
    \label{fig:city_add}
\end{center}
\end{figure}

\begin{figure}[]
\begin{center}
    \centering
    \includegraphics[width=1\textwidth]{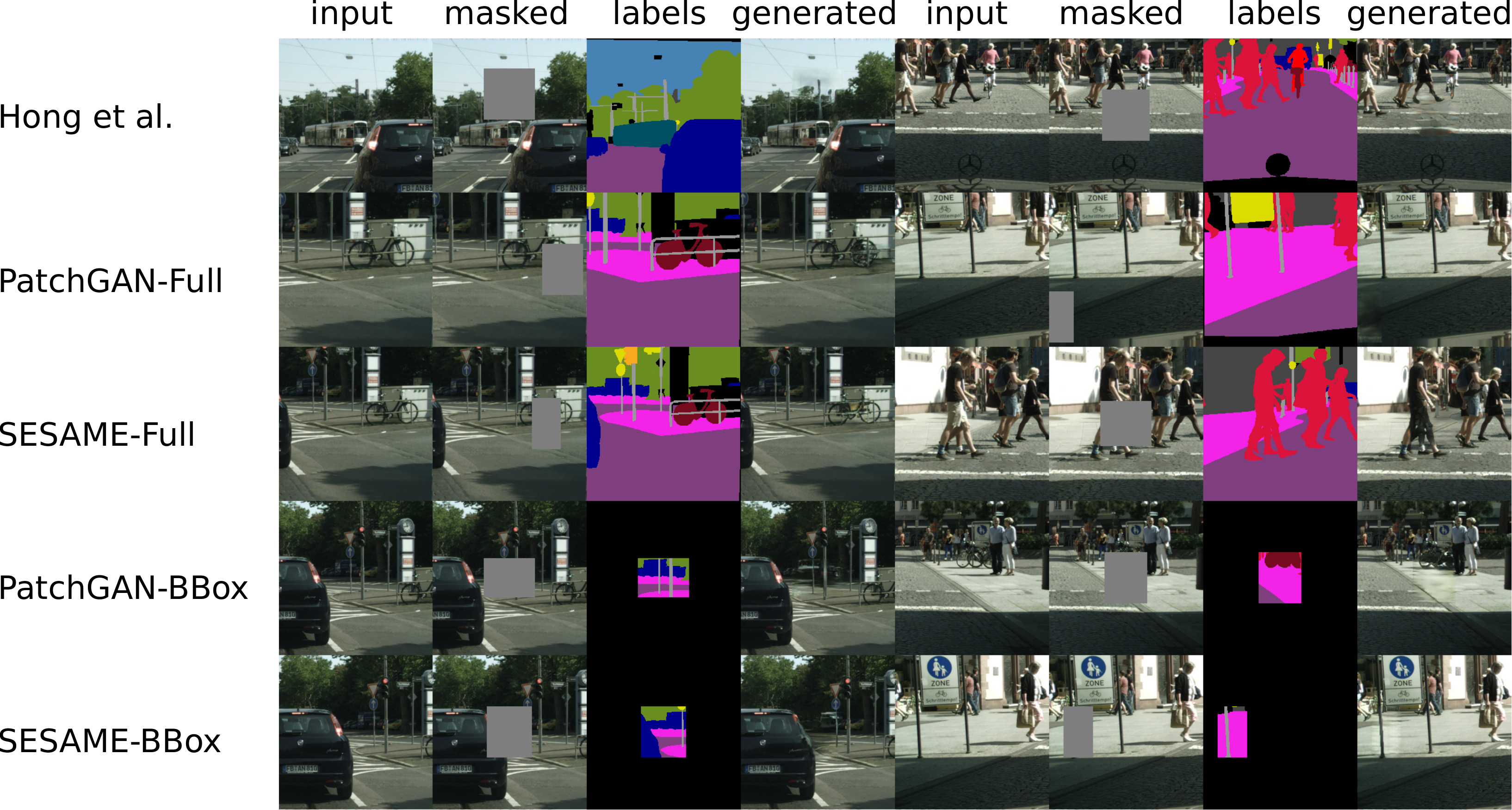}
    \caption{Visual results for removal on Cityscapes\cite{Cordts_2016_CVPR}. We ablate on: (a) using the Full context and using only the labels of the rectangular areas to be edited (b) generations due to training with the PatchGAN Discriminator and SESAME. In the first row we show the results produced by the method of \hong{}, using the full semantics information}
    \label{fig:city_remove}
\end{center}
\end{figure}

\begin{figure}[]
\begin{center}
    \centering
    \includegraphics[width=0.5\textwidth]{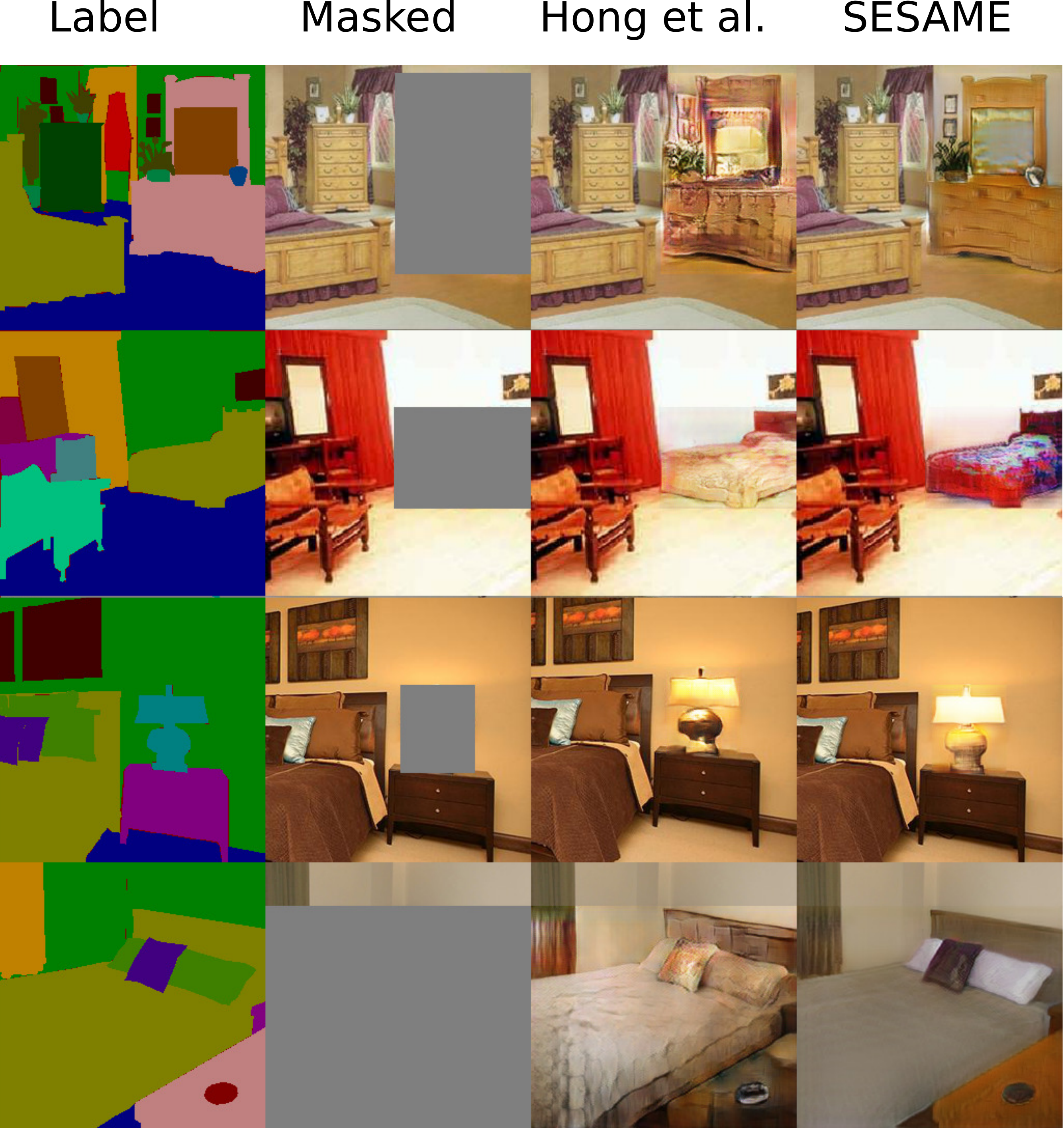}
    \caption{Visual results for editing Bedroom scenes from ADE20K dataset. Here we are using the Full semantic information to alter the gray area}
    \label{fig:ade_add}
\end{center}
\end{figure}

\begin{figure}[]
\begin{center}
    \centering
    \includegraphics[width=1\textwidth]{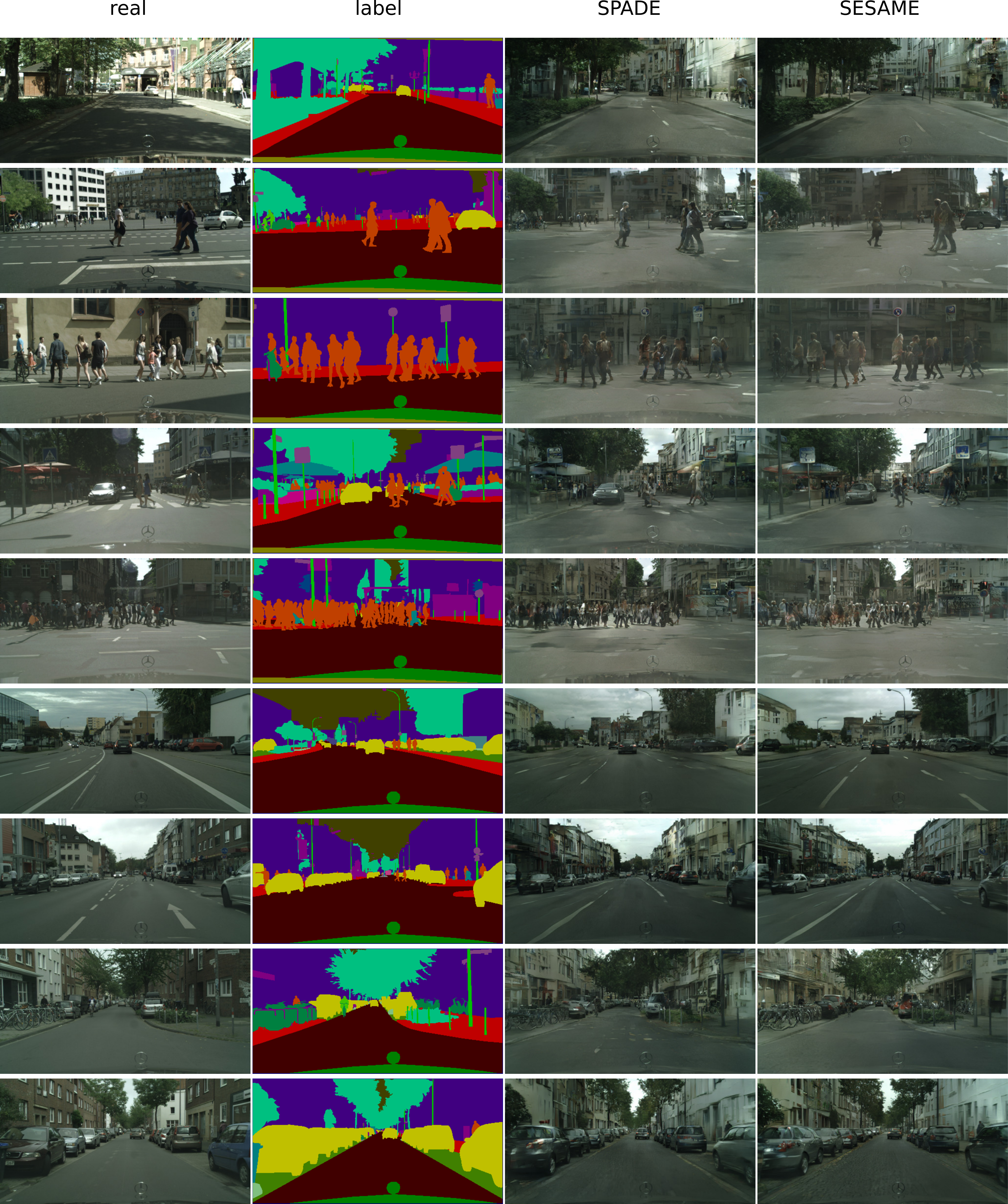}
    \caption{Cityscapes\cite{Cordts_2016_CVPR}: Visual results for image generation conditioned on Semantic Labels. We showcase the results using the generator from SPADE\cite{park2019SPADE} with the PatchGAN Discriminator(SPADE) and ours(SESAME)}
    \label{fig:city_gen}
\end{center}
\end{figure}

\begin{figure}[]
\begin{center}
    \centering
    \includegraphics[width=1\textwidth]{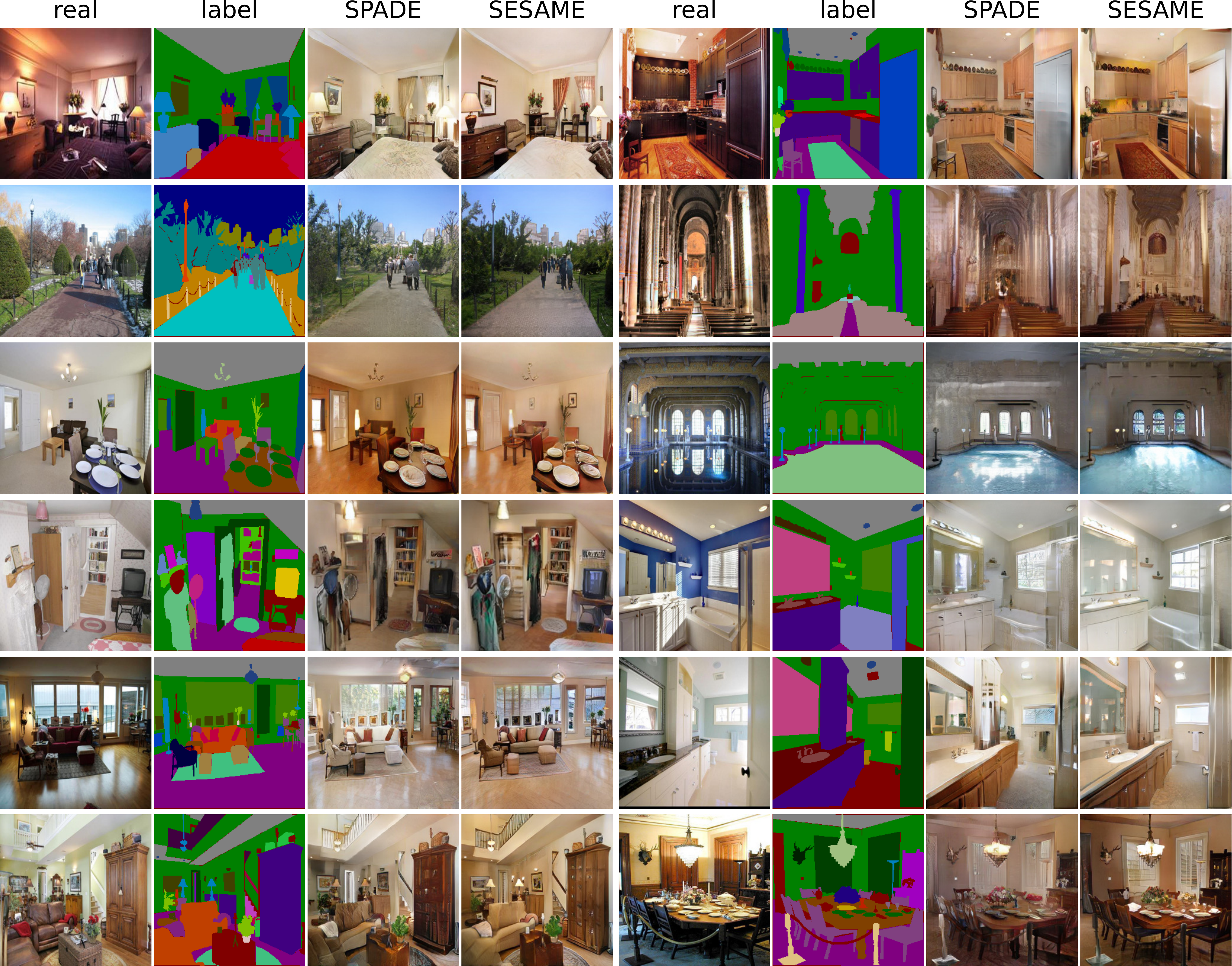}
    \caption{Ade20k\cite{zhou2017scene,zhou2016semantic}: Visual results for image generation conditioned on Semantic Labels. We showcase the results using the generator from SPADE\cite{park2019SPADE} with the PatchGAN Discriminator(SPADE) and ours(SESAME)}
    \label{fig:ade_gen}
\end{center}
\end{figure}

We show more edited and generated images produced by our method:

\begin{itemize}
    \item Figure \ref{fig:city_add} contains visual results of our ablation analysis on various access levels of semantics and different discriminators.
    \item Figure \ref{fig:city_remove} contains visual results for removing objects under different configurations.   
    \item Figure \ref{fig:ade_add} shows results for editing ADE20k-Bedroom scenes\cite{zhou2017scene,zhou2016semantic}.
    \item Figure \ref{fig:freeform} showcases examples of free-from semantic editing.   
    \item On Figures \ref{fig:city_gen} and \ref{fig:ade_gen} we can observe layout to image generation results for Cityscapes and ADE20k.   
\end{itemize}

\begin{figure}[]
\begin{center}
    \centering
    \includegraphics[width=1\textwidth]{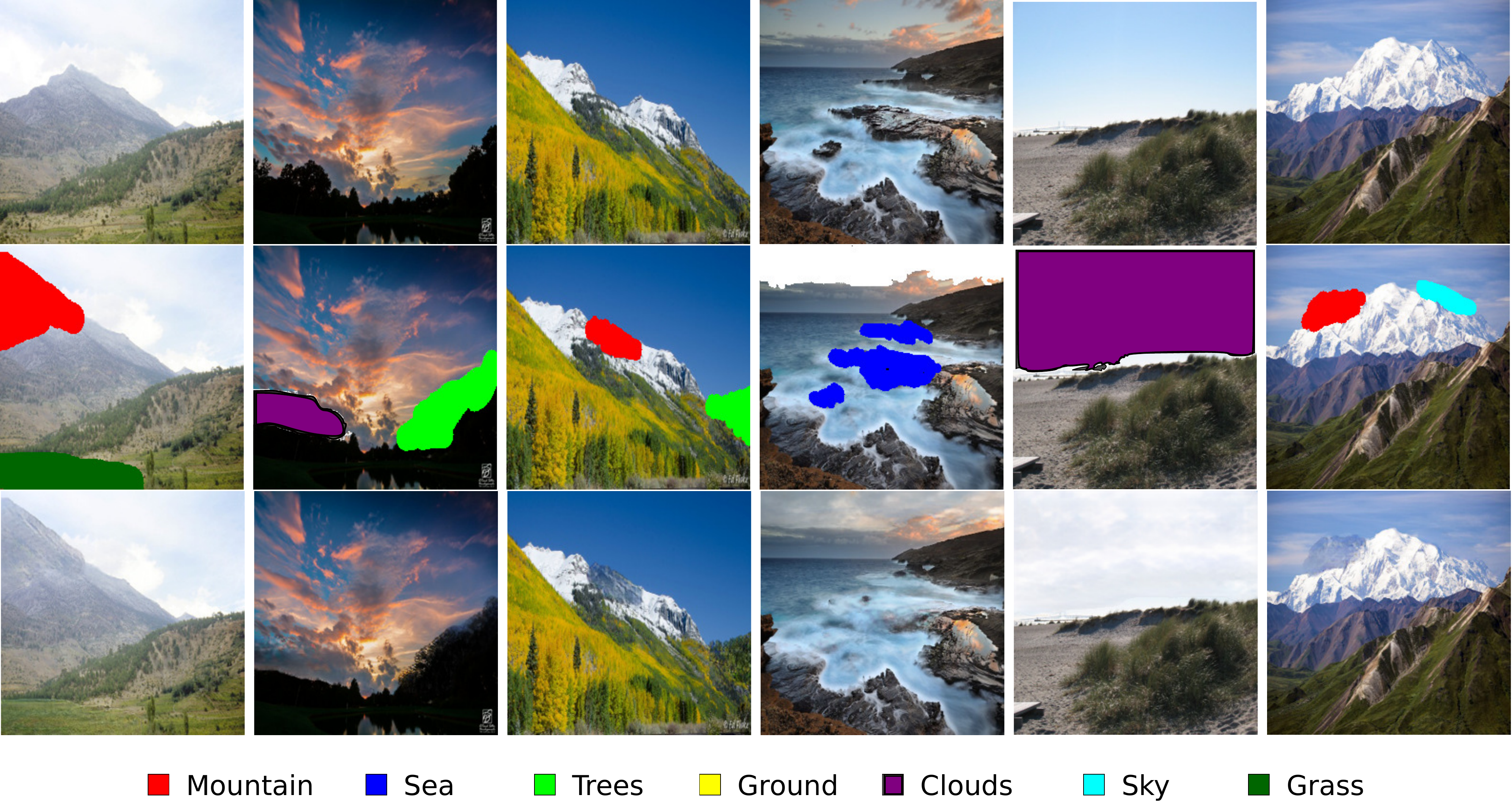}
    \caption{Examples of free form editing using semantic brushes. Note that \emph{snow} is a different semantic label from \emph{mountain} and we can observe when \textit{draw} with the mountain brush the model learned to differentiate this from snow. The model fails to correctly depict a \textit{semantic concept} out of context or with an unexpected shape}
    \label{fig:freeform}
\end{center}
\end{figure}

\subsubsection{Label to Image Generation: Comparison with CC-FPSE:}

Similarly to SESAME, \liu developed an approach to tackle image generation conditioned on semantic layouts.
They propose a generator architecture that learns to predict convolutional kernel weights conditioned on the semantic input.
Moreover, they propose a feature pyramid semantics-embedding (FPSE) discriminator using an Encoder-Decoder architecture. Each upsampling layer outputs two per-patch score maps, one trying to measure the \textit{realness} and one to gauge the \textit{semantic matching} with the labels; the later is derived after a patch-wise inner product operation with the down-sampled semantic embeddings. 

Their FPSE discriminator, while it also addresses the shortcomings of previous models, follows a different approach to our SESAME discriminator.
Although they similarly aim to short-circuit the guidance of the semantic labels to the discrimination, they choose to do so by embedding the patch with a $1\times1$ convolution and down-sampling via average pooling the semantic layout to match the size of their image processing pipeline.
As we explained in the main paper, the receptive field of a patch may contain a multitude of different semantic classes with a variety of compositions.
Trivially down-sampling the semantic label can result into loss of information.
Thus, we proposed a dedicated part of the discriminator to derive a meaningful representation for such an intricate semantic patch.
Moreover, our model independently processes the semantic information for each scale. 
 We experimented with their discriminator architecture along our SESAME generator but we were unable to achieve similar performance to our full method: SSIM 0.371, FID: 12.49, mIoU 64.24\% and accuracy 85.93\%.

\end{document}